\documentclass{article}
\usepackage[utf8]{inputenc}
\usepackage{hyperref}
\usepackage{blindtext}
\usepackage{multicol}
\usepackage{microtype}
\usepackage{graphicx}
\usepackage{subfigure}
\usepackage{booktabs}
\usepackage{svg}
\usepackage{cite}
\usepackage{amsmath,amssymb,amsfonts}
\usepackage{algorithmic}
\usepackage{graphicx}
\usepackage{textcomp}
\usepackage{xcolor}
\usepackage{color}
\usepackage{cases}
\usepackage{lipsum}
\usepackage{tabularx,booktabs}
\usepackage{tabularx,ragged2e,booktabs,caption}
\usepackage{hyperref}
\usepackage{url}
\graphicspath{{figs/}}
\newcolumntype{C}{>{\arraybackslash}X}

\usepackage{bm}
\usepackage{cleveref}
\usepackage[symbol]{footmisc}

\usepackage{footnote}
\makesavenoteenv{algorithm}

 \usepackage[accepted]{icml2021}

\newtheorem{othertheorem}{othertheorem}[section]

\newtheorem{definition}[othertheorem]{Definition}

\newcommand{\z}{\mathbf{z}}

\usepackage{array}
\newcommand{\PreserveBackslash}[1]{\let\temp=\\#1\let\\=\temp}
\newcolumntype{R}[1]{>{\PreserveBackslash\raggedleft}p{#1}}
\newcolumntype{L}[1]{>{\PreserveBackslash\raggedright}p{#1}}

\definecolor{amber}{rgb}{1.0, 0.49, 0.0}

\usepackage{color}
\usepackage{comment}
\usepackage{ulem}
\usepackage{soul}

\icmltitlerunning{Fairness in Missing Data Imputation}

\begin{document}

\twocolumn[
\icmltitle{Fairness in Missing Data Imputation}



\begin{icmlauthorlist}
\icmlauthor{Yiliang Zhang}{to}
\icmlauthor{Qi Long}{med}
\end{icmlauthorlist}

\icmlaffiliation{to}{Department of Applied Mathematics and Computational Science, University of Pennsylvania, United States}
\icmlaffiliation{med}{Department of Biostatistics, Epidemiology and Informatics, University of Pennsylvania, United States}
\icmlcorrespondingauthor{Qi Long}{qlong@pennmedicine.upenn.edu}

\icmlkeywords{Missing value, Imputation, Semi-supervised Learning, ICML}
\vskip 0.3in
]

\printAffiliationsAndNotice{}


\begin{abstract}
Missing data are ubiquitous in the era of big data and, if inadequately handled, are known to lead to biased findings and have deleterious impact on data-driven decision makings.  To mitigate its impact, many missing value imputation methods have been developed. However, the fairness of these imputation methods across sensitive groups has not been studied. In this paper, we conduct the first known research on fairness of missing data imputation. By studying the performance of imputation methods in three commonly used datasets, we demonstrate that unfairness of missing value imputation widely exists and may be associated with multiple factors. Our results suggest that, in practice, a careful investigation of related factors can provide valuable insights on mitigating unfairness associated with missing data imputation.
\end{abstract}
\vspace{-0.7cm}

\section{Introduction}

Missing data, as well-recognized, have significant impact on analysis of real data in many fields. One of the most popular approaches for handling missing data is missing data imputation. With the development of statistics and machine learning, different approaches have been adopted for the imputation task \citep{van1999flexible,stekhoven2012missforest,gondara2018mida, li2019misgan, yoon2018gain, yang2019categorical}. While these machine learning imputation methods show preferable performances, potential concern about unfairness of the algorithms is non-negligible. Machine learning algorithms have been shown to inherit the bias and unfairness that human have in decision making. Various studies are conducted in the context of computer vision and natural language processing \citep{buolamwini2018gender, klare2012face, ryu2017inclusivefacenet, bolukbasi2016man,brunet2019understanding, gonen2019lipstick, zhao2018gender, zhao2019gender, zhang2020fairness}, illustrating the wide existence of unfairness and possible remedies. Algorithmic fairness, focusing on non-discrimination of decision outcomes, comes to the fore in the research community.

As discussed in \citet{chouldechova2018frontiers}, \citet{martinez2019fairness} and \citet{pessach2020algorithmic}, the causes of unfairness in machine learning mainly come from bias in dataset, missing values, prediction algorithms as well as imbalance of populations among different sensitive groups. In particular, it's discussed in \citet{bakker2020fair}, \citet{martinez2019fairness}, \citet{rajkomar2018ensuring} and \citet{gianfrancesco2018potential} that missing values contribute to the bias of algorithms. Empirical analysis in \citet{martinez2019fairness}  brings to the fore the issue of correctly handling missing data in the sense of fairness, instead of dropping out corresponding samples directly. Inspired by this work, we study the fairness of some representative imputation methods in three datasets\footnote{A detailed description of the datasets can be found in \Cref{ap:data}}: COMPAS recidivism, Alzheimer’s Disease Neuroimaging Initiative (ADNI) and atherosclerosis cardiovascular disease (ASCVD). By looking into imputation results and corresponding predictions for each imputation method and missing data mechanism, we observe non-negligible bias among different gender and race groups among all the imputation methods compared.


In the context of the impact missing data have on fairness, \citet{martinez2019fairness} studies fairness of existing real datasets with missing values; \citet{wang2021analyzing} studies the impact of missing values in categorical data on fairness; \citet{goel2020importance} illustrates the recoverability of outcome's distribution from complete cases, through a causal modeling of data missingness. Our work is the first work to systematically study the fairness associated with the process of imputing missing data.


\textbf{Our contributions:} In this paper, we propose the first fairness notion in imputation, \textit{imputation accuracy parity difference}. We also study fairness of prediction model built on imputated datasets. Our empirical studies show that (1) Severe unfairness exists in both imputation and prediction after imputation. (2) Different imputation methods have non-negligible impact on fairness. (3) Unfairness in both imputation and prediction can be associated with the sample imbalance and missing data mechanism. Our work provides valuable insight into how to effectively handle missing values while guaranteeing fairness of the learning tasks.

\vspace{-0.13in}
\section{Preliminaries}
\vspace{-0.07in}

\subsection{Fairness notions}
\vspace{-0.05in}

Discussion on fairness are based on pre-specified sensitive attributes $A$. In real data experiments, we use gender and ethnicity as sensitive attributes, which are widely used in social science study. For gender, we identify male as the majority group and female as the minority group. For ethnicity we identify white as the majority group and black (or ``other races'') as the minority group. Regarding the fairness notions, Equalized Odds (EO) \citep{hardt2016equality} is widely used. For a decision making procedure, EO aims to let algorithm's prediction $\hat{y}$ independent of sensitive attribute $A$, conditioned on the true outcome $y$. Consider the learning task in which data $\mathbf{Z}$ contains predictor $X$ and response $y$. When response $y$ is binary, false positive rate (FPR) and false negative rate (FNR) for classifier $h$ in group $a$ are defined as $\text{FPR}_{a}(h) := \sum_{A_i = a}\mathbf{1}\{h(x_i) = 1\}/\sum_{A_i = a}\mathbf{1}\{y_i = 0\}$ and $\text{FNR}_{a}(h) := \sum_{A_i = a}\mathbf{1}\{h(x_i) = 0\}/\sum_{A_i = a}\mathbf{1}\{y_i = 1\}$. We define \textit{equalized odds difference} as the fairness notion in classification:
\vspace{-0.2cm}

\begin{definition}[equalized odds difference]\label{eod}
Equalized odds difference for classifier $h$ is defined as $\text{EOD}(h) := |\text{FPR}_{\text{maj}}(h) - \text{FPR}_{\text{min}}(h)| + |\text{FNR}_{\text{maj}}(h) - \text{FNR}_{\text{min}}(h)|$.
\end{definition}
\vspace{-0.2cm}

Here the subscripts ``maj'' and ``min'' mean the quantities are for majority and minority sensitive groups, respectively. The notion is a measure of how close the prediction algorithm $h$ is to the equalized odds, in which EOD equals to 0.


In addition, in this paper we propose a novel notion that measures the fairness of imputation results. We narrow our scope to the scenario when sensitive attributes are binary $A = a \in \{0,1\}$. Suppose the complete data matrix, without missing values, is denoted by $\mathbf{Z} = (z_{ij}) \in \mathbb{R}^{n \times p}$, and the missing indicator is denoted by $\mathbf{R} = (r_{ij}) \in \mathbb{R}^{n \times p}$, with $r_{ij} = \mathbf{1}\{z_{ij} \text{ is observed}\}$. Let $\mathbf{Z}_{\text{obs}} = \{z_{ij}| r_{ij} = 1\}$ denote the observed data and $\mathbf{Z}_{\text{miss}} = \{z_{ij}| r_{ij} = 0\}$ denote the missing data. We further let $\mathbf{Z}^a = (z^a_{ij})$ and $\mathbf{R}^a = (r^a_{ij})$ denote the complete data matrix and missing indicator matrix in sensitive group $A = a$ (so that $\mathbf{Z} = \mathbf{Z}^0 \bigcup \mathbf{Z}^1$, $\mathbf{R} = \mathbf{R}^0 \bigcup \mathbf{R}^1$). In group $a$, the data matrix imputed by model $g$ is denoted by $\widehat{\mathbf{Z}}^a(g) = (\hat{z}^a_{ij}(g))$. Assume that both sensitive groups contain missing data, we define mean square imputation error of $g$ in group $a$ as
$$
\text{MSIE}_a(g) = \frac{\sum_{(i,j)}(\hat{z}^a_{ij}(g) - {z}^a_{ij})^2(1-r^a_{ij})}{\sum_{(i,j)} 1-r^a_{ij}}
$$
Now we define \textit{imputation accuracy parity}, a novel notion that measures fairness of imputation model $g$:



\begin{table*}[t]
\scalebox{0.8}{
\setlength\tabcolsep{2pt}
\begin{tabular}{c|cccccccc|cccccccc|cccccccc}
\toprule
     & \multicolumn{8}{c}{MSIE} & \multicolumn{8}{c@{}}{IAPD regarding gender} & \multicolumn{8}{c@{}}{IAPD regarding race}\\
    \midrule
     Method & $I_1$ & $I_2$ & $I_3$ & $I_4$ & $I_5$ & $I_6$ & $I_7$ & Var &
     $I_1$ & $I_2$ & $I_3$ & $I_4$ & $I_5$ & $I_6$ & $I_7$ & VarD$_g$ &
     $I_1$ & $I_2$ & $I_3$ & $I_4$ & $I_5$ & $I_6$ & $I_7$ & VarD$_r$ \\
\midrule
\addlinespace

MCAR (\ref{mcar-1})   
 & 0.68  & 0.68  & \textbf{0.67} & 0.81 & 0.88 & 0.73 & 0.99 & 0.99
 & 0.45  & 0.46  & 0.47 & 0.50 & 0.48 & \textbf{0.36} & 0.51 & 0.55 
 & -0.46  & -0.48  & -0.48 & -0.49 & -0.53 & \textbf{-0.32} & -0.50 & -0.53 
  \\
MCAR (\ref{mcar-2})   
 & 0.90  & 0.82  & \textbf{0.80} & 0.91 & 0.87 & 0.86 & 0.95 & 1.00
 & 0.58  & 0.51  & 0.51 & 0.52 & 0.50 & \textbf{0.42} & 0.52 & 0.56
 & -0.57  & -0.56  & -0.55 & -0.52 & -0.54 & \textbf{-0.42} & -0.49 & -0.54
  \\

MCAR (\ref{mcar-3})   
 & \textbf{0.85}  & 1.09  & 0.88 & 0.96 & 0.88 & 0.95 & 0.95 & 1.00
 & 0.54  & 0.58  & 0.53 & 0.53 & 0.51 & \textbf{0.44} & 0.47 & 0.56
 & -0.55  & -0.73  & -0.59 & -0.53 & -0.55 & \textbf{-0.43} & -0.46 & -0.54
  \\ 

\addlinespace
MAR (\ref{mar-1})   
 & \textbf{0.89}  & 1.08  & 0.85 & 1.03 & 0.90 & 1.00 & 1.04 & 1.09
 & 0.53  & 0.79  & 0.57 & 0.56 & \textbf{0.50} & 0.60 & \textbf{0.50} & 0.55 
 & -0.63  & -0.76  & -0.64 & -0.61 & -0.62 & \textbf{-0.51} & -0.57 & -0.61
  \\ 
  
MAR (\ref{mar-2})   
 & \textbf{0.47}  & 0.53  & 0.49 & 0.62 & 0.57 & 0.70 & 0.69 & 0.72
 & 0.43  & 0.35  & 0.41 & 0.49 & 0.49 & \textbf{0.14} & 0.51 & 0.55 
 & -0.23  & -0.27  & -0.25 & -0.24 & -0.28 & \textbf{-0.15} & -0.22 & -0.27
  \\

MAR (\ref{mar-3})   
 & 0.42  & \textbf{0.36}  & 0.54 & 0.51 & 0.53 & 0.80 & 0.62 & 0.57
 & 0.15  & 0.15  & 0.13 & 0.18 & 0.19 & \textbf{0.11} & 0.17 & 0.21 
 & -0.12  & -0.22  & \textbf{-0.04} & -0.21 & -0.23 & \textbf{-0.04} & -0.06 & -0.25
  \\

MAR (\ref{mar-4})    
 & \textbf{1.27} & 1.66 & 1.35  & 1.36 & 1.34 & 1.37 & 1.48 & 1.44
 & 0.86  & 0.85  & 0.92 & 0.83 & 0.79 & \textbf{0.73} & 0.84 & 0.83 
 & -0.72  & -0.87  & -0.81 & -0.64 & \textbf{-0.60} & -0.65 & -0.77 & -0.62
  \\

\addlinespace
MNAR (\ref{mnar-1})   
 & 1.08  & 1.16  & 2.44 & \textbf{0.26} & 0.36 & 0.65 & 0.64 & 0.12
 & 0.23  & 0.21  & 0.14 & \textbf{0.00} & -0.01 & \textbf{0.00} & 0.01 & 0.01
 & -0.35  & -0.47  & -0.30 & 0.03 & 0.02 & \textbf{0.01} & 0.08 & 0.02
  \\

MNAR (\ref{mnar-2})   
 & 0.67  & 0.53  & 0.67 & \textbf{0.28} & 0.35 & 0.60 & 0.52 & 0.27
 & 0.25  & 0.16  & 0.14 & 0.02 & \textbf{0.01} & \textbf{-0.01} & 0.03 & 0.05 
 & -0.27  & -0.32  & -0.29 & \textbf{0.00} & 0.03 & 0.02 & -0.05 & -0.03
  \\

MNAR (\ref{mnar-3})   
 & 1.50  & \textbf{1.42} & 1.62 & 1.59 & 1.43 & 1.67 & 1.85 & 1.59
 & 1.09  & 1.06  & 1.14 & 1.04 & 1.00 & \textbf{0.98} & 1.04 & 0.95 
 & -1.01  & -1.02  & -1.13 & -1.01 & -1.01 & \textbf{-0.87} & -0.97 & -0.89
  \\

MNAR (\ref{mnar-4})   
 & 2.71  & 2.44  & 2.53 & 2.20 & \textbf{1.88} & 1.95 & 2.11 & 1.71
 & 1.59  & 1.47  & 1.56 & 1.35 & 1.28 & \textbf{1.01} & 1.03 & 1.00 
 & -1.18  & -1.32  & -1.25 & -1.20 & -1.20 & \textbf{-0.83} & -0.88 & -0.90
  \\

\addlinespace
\bottomrule
\end{tabular}
}
\caption{Imputation fairness on COMPAS recidivism dataset. Number of features (besides sensitive attributes) is 10, $L$ = 5. Here imputation methods are encoded as: $I_1$: MICE; $I_2$: missForest; $I_3$: KNN; $I_4$: SoftImpute; $I_5$: OptSpace; $I_6$: Gain; $I_7$: misGAN. Results are average values over 50 repeated experiments. Var denotes the variance of missing data, VarD$_g$ denotes the difference of missing value's variances between two gender groups and VarD$_r$ denotes that between two race groups.}
\vspace{-0.1in}
\label{compas_imp}
\end{table*}
\normalsize

\begin{definition}[imputation accuracy parity difference]
Imputation accuracy parity for imputation model $g$ is defined as $\text{IAPD}(g) = \text{MSIE}_{\text{maj}}(g) - \text{MSIE}_{\text{min}}(g)$. 
\end{definition}
\vspace{-0.1in}
Imputation accuracy parity is similar to the fairness notion \textit{accuracy parity} adopted in multiple literature \citep{zafar2017fairness,friedler2016possibility,zhao2019conditional}. Consider the learning tasks where data $\mathbf{Z}$ contains predictor $X$ and binary response $y$. When only $y$ contains missing values, imputation can be regarded as a prediction task. In such case, accuracy parity for imputation method $g$ can be regarded as a finite-sample version of the accuracy parity.
\vspace{-0.1in}
\subsection{Missing data mechanism}
\vspace{-0.1in}
The missing data mechanism \citep{little2019statistical} can be classified into three types: missing completely at random (MCAR), missing at random (MAR) and missing not at random (MNAR). A missing mechanism is said to be MCAR if missingness is independent of both observed and missing data. When missingness is only dependent on observed data, the mechanism is said to be MAR. For the MNAR, missingness can be associated with both observed and unobserved data. In this paper, we conduct real data experiments to assess fairness associated with imputation in which missing data are artificially generated under all three mechanisms. Specifically, for each real data set used, we normalize all the features and then generate missing values in the first $L$ features in the dataset, with a pre-specified $L$. Throughout, we use the following 11 models for generating missing values. Given a sample $\z = (z_1, \cdots, z_p)$, the probability that $z_{j}$ is missing is given by the following values\footnote{the values are truncated inside the unit interval $[0,1]$}, for $\forall j \in \{1,\dots,L\}$:

\noindent
\begin{minipage}[t]{0.2\linewidth}
\centering
\paragraph{MCAR}
\small
\vspace{-0.05in}
\begin{subequations}
\begin{align}
&0.1 \label{mcar-1}
   \\
   &0.5 \label{mcar-2}
   \\
   &0.9 \label{mcar-3}
\end{align}
\end{subequations}

\normalsize
\end{minipage} 
\begin{minipage}[t]{0.38\linewidth}
\centering
\paragraph{MAR}
\small
\vspace{-0.05in}
\begin{subequations}
\begin{align}
   &0.1 + 0.8 \mathbf{1}_{\text{male}} \label{mar-1}
   \\
   &0.1 + 0.8 \mathbf{1}_{\text{female}} \label{mar-2}
   \\
   &0.5 - 0.5 z_{L+j} \label{mar-3}
   \\
   &0.5 + 0.5 z_{L+j} \label{mar-4}
\end{align}
\end{subequations}
\normalsize
\end{minipage}
\begin{minipage}[t]{0.38\linewidth}
\centering
\paragraph{MNAR}
\small
\vspace{-0.05in}
\begin{subequations}
\begin{align}
   &0.5 - z_{j} \label{mnar-1}
   \\
   &0.5 - 0.2 z_{j} \label{mnar-2}
   \\
   &0.5 + 0.2 z_{j} \label{mnar-3}
   \\
   &0.5 + z_{j} \label{mnar-4}
\end{align}
\end{subequations}
\normalsize
\end{minipage}

\subsection{Imputation methods}
The imputation methods investigated include MICE \citep{buuren2010mice}, missForest \citep{stekhoven2012missforest}, K-nearest neighbor (KNN) imputation, two matrix completion methods SoftImpute \citep{hastie2015matrix} and OptSpace \citep{keshavan2010matrix}, and also two deep learning methods Gain \citep{yoon2018gain} and Misgan \citep{li2019misgan}. A more thorough review on existing imputation methods is provided in \Cref{ap:alg}.





\vspace{-0.1in}
\section{Fairness in imputation accuracy}
\vspace{-0.1in}

In this section, we investigate the imputation fairness for different existing methods under various missing mechanisms. Each experiment is repeated for 50 times and results shown are the average values. Let VarD$_g$ and VarD$_r$ denote the difference of missing value's variances between two gender groups and between two racial groups, respectively. Both of the quantities serve as baselines of imputation performance, since the variance difference is equivalent to IAPD when using imputing missing values with mean observed values in two groups respectively. We set $L = 5$ in the experiment of COMPAS dataset and report the results in Table (\ref{compas_imp}). Results for two other real datasets are provided in \Cref{adni} and \ref{cvd}.



\noindent \textbf{Observation 1: Severe imputation unfairness widely exists}

We observe that in all the 11 missing mechanisms, almost all the imputation methods have positive imputation IAPD regarding gender and negative IAPD regarding race. This implies that regarding gender, all the imputation methods consistently provide more accurate imputation result for female group compared with male group. Meanwhile, all the imputation methods give more accurate imputation result for white people compared with black people. Severe imputation unfairness are observed in this dataset among all the imputation models. In \Cref{adni} and \ref{cvd}, we also observe imputation unfairness in ADNI and ASCVD datasets.

\noindent \textbf{Observation 2: Imputation fairness can be influenced by imbalance of missingness}


The missingness in MAR (\ref{mar-1}) and (\ref{mar-2}) are approximately 0.5, which is in the similar level as MCAR (\ref{mcar-2}). However, in mechanism (\ref{mar-1}), about 90$\%$ of first 5 features in male group are missing while only 10$\%$ of that in female group are missing. This difference can cause significant influence on imputation fairness: The more missingness one group has, the larger imputation error it appears to have. From Table \ref{compas_imp}, for many imputation methods we found that imputation IAPD regarding gender in MAR (\ref{mar-1}) is consistently larger (taking sign into account) than that in mechanism MCAR (\ref{mcar-2}). This indicates male group's imputation error becomes relatively larger in MAR (\ref{mar-1}). Meanwhile, all the imputation methods gives smaller imputation IAPD regarding gender in MAR (\ref{mar-2}) than that in mechanism MCAR (\ref{mcar-2}). These empirical observation indicates that for a fixed overall missingness, imbalance of missingness between two sensitive groups can also influence imputation fairness effectively. Additional evidence is also observed in the experiments of ADNI dataset, shown in \Cref{adni}.


\noindent \textbf{Observation 3: Imputation unfairness tends to be enlarged as missingness increases}

Among 3 MCAR mechanisms, from (1a) to (1c) the missingness increases (from 0.1 to 0.9). A trend of increasing imputation unfairness (IAPD regarding gender and race) is also observed for most imputation methods. This implies that degree of missingness contributes to the imputation fairness. Intuitively, the bias in imputation can be amplified by increasing missingness.




\begin{table*}[t]
\scalebox{0.8}{
\setlength\tabcolsep{2.6pt}
\begin{tabular}{c|cccccccc|cccccccc|cccccccc}
\toprule
     & \multicolumn{8}{c}{Prediction Accuracy} & \multicolumn{8}{c@{}}{EOD regarding gender} & \multicolumn{8}{c@{}}{EOD regarding race}\\
    \midrule
     Method & $I_1$ & $I_2$ & $I_3$ & $I_4$ & $I_5$ & $I_6$ & $I_7$ & CC &
     $I_1$ & $I_2$ & $I_3$ & $I_4$ & $I_5$ & $I_6$ & $I_7$ & CC$_g$ &
     $I_1$ & $I_2$ & $I_3$ & $I_4$ & $I_5$ & $I_6$ & $I_7$ & CC$_r$ \\
\midrule
\addlinespace

MCAR (\ref{mcar-1})   
 & 0.71  & 0.71  & 0.71 & 0.63 & 0.65 & 0.60 & 0.71 & \textbf{0.72}
 & 0.18  & 0.16  & 0.16 & \textbf{0.05} & 0.06 & 0.06 & 0.17 & 0.16 
 & 0.29  & 0.29  & 0.29 & \textbf{0.03} & 0.05 & 0.04 & 0.30 & 0.29
  \\
  
MCAR (\ref{mcar-2})   
 & 0.68  & 0.70  & 0.68 & 0.63 & 0.65 & 0.58 & 0.70 & \textbf{0.71}
 & 0.17  & 0.16  & 0.14 & 0.05 & 0.06 & \textbf{0.03} & 0.17 & 0.16
 & 0.27  & 0.29  & 0.26 & \textbf{0.03} & 0.04 & 0.05 & 0.31 & 0.29
  \\

MCAR (\ref{mcar-3})   
 & 0.67  & 0.66  & 0.66 & 0.64 & 0.65 & 0.64 & 0.64 & \textbf{0.70}
 & 0.17  & 0.13  & 0.13 & \textbf{0.06} & \textbf{0.06} & 0.08 & 0.12 & 0.15
 & 0.23  & 0.24  & 0.23 & \textbf{0.04} & 0.07 & 0.15 & 0.20 & 0.29
  \\ 

\addlinespace
MAR (\ref{mar-1})   
 & 0.69  & 0.69  & 0.69 & 0.64 & 0.65 & 0.63 & 0.69 & \textbf{0.70}
 & 0.14  & 0.15  & 0.13 & 0.06 & \textbf{0.05} & 0.06 & 0.14 & 0.26
 & 0.24  & 0.24  & 0.25 & \textbf{0.03} & 0.04 & 0.11 & 0.24 & 0.27
  \\ 
  
MAR (\ref{mar-2})   
 & 0.70  & 0.70  & \textbf{0.71} & 0.64 & 0.65 & 0.61 & 0.70 & \textbf{0.71}
 & 0.17  & 0.15  & 0.16 & 0.06 & \textbf{0.05} & \textbf{0.05} & 0.17 & 0.16
 & 0.31  & 0.30  & 0.29 & 0.04 & \textbf{0.03} & 0.04 & 0.30 & 0.29
  \\

MAR (\ref{mar-3})   
 & 0.46  & 0.46  & 0.46 & 0.64 & 0.65 & 0.52 & 0.46 & \textbf{0.71}
 & 0.01  & \textbf{0.00}  & 0.02 & 0.06 & 0.06 & 0.03 & 0.03 & 0.16 
 & 0.02  & \textbf{0.00}  & 0.02 & 0.03 & 0.07 & 0.15 & 0.06 & 0.30
  \\

MAR (\ref{mar-4})    
 & 0.57 & 0.61 & 0.59  & 0.54 & 0.65 & 0.50 & 0.54 & \textbf{0.71}
 & 0.01  & 0.06  & 0.03 & \textbf{0.00} & 0.07 & 0.01 & 0.01 & 0.17
 & \textbf{0.00} & 0.05  & 0.02 & \textbf{0.00} & 0.05 & 0.10 & 0.03 & 0.29
  \\

\addlinespace
MNAR (\ref{mnar-1})   
 & 0.68  & 0.66  & 0.62 & 0.64 & 0.66 & 0.61 & 0.64 & \textbf{0.69}
 & 0.12  & 0.12  & 0.09 & 0.05 & 0.06 & 0.04 & \textbf{0.01} & 0.14
 & 0.32  & 0.24  & 0.16 & \textbf{0.03} & 0.35 & \textbf{0.03} & 0.08 & 0.25
  \\

MNAR (\ref{mnar-2})   
 & 0.67  & 0.69  & 0.69 & 0.64 & 0.65 & 0.62 & 0.67 & \textbf{0.71}
 & 0.12  & 0.16  & 0.16 & \textbf{0.05} & 0.08 & \textbf{0.05} & 0.06 & 0.16 
 & 0.26  & 0.28  & 0.26 & \textbf{0.03} & 0.17 & 0.04 & 0.16 & 0.29
  \\

MNAR (\ref{mnar-3})   
 & 0.70  & 0.69 & 0.68 & 0.64 & 0.65 & 0.57 & 0.69 & \textbf{0.71}
 & 0.17  & 0.13  & 0.12 & 0.06 & 0.06 & \textbf{0.03} & 0.09 & 0.16 
 & 0.31  & 0.26  & 0.25 & 0.04 & \textbf{0.03} & 0.05 & 0.18 & 0.28
  \\

MNAR (\ref{mnar-4})   
 & 0.69  & 0.68  & 0.69 & 0.64 & 0.65 & 0.58 & 0.69 & \textbf{0.71}
 & 0.12  & 0.09  & 0.09 & 0.07 & 0.07 & \textbf{0.03} & 0.15 & 0.16 
 & 0.27  & 0.26  & 0.26 & \textbf{0.04} & 0.06 & 0.05 & 0.31 & 0.28
  \\

\addlinespace
\bottomrule
\end{tabular}
}
\caption{Prediction fairness on COMPAS recidivism dataset. Number of features (besides sensitive attributes) is 10, $L$ = 5. Here imputation methods are encoded as: $I_1$: MICE; $I_2$: missForest; $I_3$: KNN; $I_4$: SoftImpute; $I_5$: OptSpace; $I_6$: Gain; $I_7$: misGAN. CC: prediction using complete cases in the training set. Results are average values over 50 repeated experiments. When using complete data for prediction, prediction accuracy is 0.72, accuracy difference regarding gender is 0.16, accuracy difference regarding race is 0.29. }
\vspace{-0.15in}
\label{compas_pred}
\end{table*}
\normalsize

Besides, in \Cref{adni} we have an additional observation that imputation unfairness can be associated with imbalance of sample size. Of all 649 patients in ADNI gene dataset, 642 patients are white people. We observe that the IAPD regarding race groups is almost always negative, implying imputation methods' preferences towards white people during imputation. This effect of sample imbalance is also observed in ASCVD dataset (\Cref{cvd}), where population in white group is two times that in the other group and that IAPD regarding race groups is also consistently negative. This observation matches intuition that imputation accuracy is positively correlated with the amount of information observed. Imbalance of observed information leads to imputation unfairness.

\vspace{-0.1in}
\section{Fairness in prediction accuracy}
\vspace{-0.05in}

We study the prediction fairness for different imputation methods on three aforementioned datasets, each of which contains a response $y$. In each experiment, we firstly conduct a train-test split (80 $\%$ and 20 $\%$ of the sample size, respectively), artificially generate missing data in the training set, according to the 11 missing mechanisms in Section 2. Next, we impute the missing data using all the imputation models mentioned in Section 2.3. Finally, we train a prediction model $h$ using random forest, based on the original training set (without missing values), complete cases in the training set and imputed datasets. Equalized odds difference (Definition \ref{eod}) on the test set are reported. The results for COMPAS is shown in Table \ref{compas_pred} and those for other two datasets are provided in \Cref{adni} and \ref{cvd}.




\noindent \textbf{Observation 4: Prediction fairness is associated with missing mechanism}

We observe that for each imputation method, i.e., a fixed column, the EOD values vary cross different rows. This implies that different missing mechanisms lead to different prediction fairness when the imputation method is fixed. In particular, in MAR (\ref{mar-3}), all the imputation methods has smaller EOD regarding both gender and race, compared with the EODs when using complete data or complete cases to build the prediction model.



\noindent \textbf{Observation 5: Imputation posts a trade-off between accuracy and fairness in prediction}

From Table \ref{compas_pred}, we observe that for a fixed missing mechanism, prediction fairness associated with different imputation methods are different. In particular, we observe that Gain and two matrix completion methods: SoftImpute and OptSpace consistently have smaller EOD compared with other imputation methods (and prediction models without imputation, using complete cases and complete data).

Meanwhile, prediction algorithms associated with these three methods (i.e., Gain, SoftImput and OptSpace) have lower prediction accuracy. This can be viewed as a trade-off between prediction accuracy and prediction fairness. In fact, we notice from the table that such trade-off widely exists. For an arbitrary missing mechanism, the prediction accuracy associated with an imputation method is lower than that associated with complete data, and most imputation models are also associated with a smaller EOD compared with that associated with the morel built through the complete data. In addition, in the experiment of ADNI data (shown in Table \ref{adni_pred}), the EOD for race is larger than 0.15 in most cases. A potential reason is that 624 out of 649 samples are from white patients. This suggests that prediction unfairness can also be influenced by the sample imbalance in some extreme cases, including our experiments of ADNI data.




\vspace{-0.1in}
\section{Discussion}
\vspace{-0.05in}

In this paper we study the fairness associated with missing data imputation in three real datasets. Our experiments show that imputation unfairness widely exists among different imputation models, representing the first known empirical results in literature. We also demonstrate  factors that could contribute to imputation fairness. We further study the impact of imputation on prediction fairness when imputed data are used to build prediction models. This area offers fertile ground for theoretical investigation, as there has been littler exiting work in this area. Lastly, we acknowledge the ADNI for providing the gene datasets. \footnote{ \url{http://adni.loni.usc.edu/wp-content/uploads/how_to_apply/ADNI_Acknowledgement_List.pdf}.}

\vspace{-0.12in}
\section*{Acknowledgements}
\vspace{-0.1in}
This work is partly supported by an NIH grant R01GM124111.

\newpage

\bibliography{ref}
\bibliographystyle{icml2021}
\clearpage
\appendix

\section{Datasets}\label{ap:data}

Among three real datasets analyzed in the paper, COMPAS recidivism dataset has been widely used in fairness literature. The rest two datasets are high-dimensional, which is also in our interest. Throughout the paper, all the data are normalized in a column-wise manner.

\paragraph{COMPAS Recidivism}

COMPAS (Correctional Offender Management Profiling for Alternative Sanctions) \cite{COMPAS} is a risk assessment instrument developed by Northpointe Inc., which predicts recidivism of defendants. The dataset contains demographic information and criminal history of each defendants, along with his recidivism in the following two years. Existence of certain bias has been illustrated in COMPAS algorithm towards certain group of defendants (e.g. race, gender and age) \cite{angwin2016compas}. The dataset has been widely used in existing fairness literature \cite{corbett2017algorithmic, grgic2016case, wadsworth2018achieving, zafar2017fairness}. We take all the 10 numerical features out of total 21 features and let the age, prior crime counts and other 3 non-categorical variables to have missing values.

\paragraph{ADNI gene expression}


The data were obtained from the Alzheimer’s Disease Neuroimaging Initiative (ADNI) database (\url{adni.loni.usc.edu}). The ADNI was launched in 2003 as a public-private partnership, led by Principal Investigator Michael W. Weiner, MD. The primary goal of ADNI has been to test whether serial magnetic resonance imaging (MRI), positron emission tomography (PET), other biological markers, and clinical and neuropsychological assessment can be combined to measure the progression of mild cognitive impairment and early Alzheimer’s disease. The dataset contains information of 649 patient's gene expression, who is potentially with Alzheimer disease. More than 19k features are recorded in the original dataset and for the sake of computation efficiency, we only select 1000 features from the raw data in experiment, which have the largest positive correlations with gender (encode `male' as 1 and `female' as 0). The outcome variable for prediction is disease status, containing four categories: cognitively normal (CN), two types of mild cognitive impairments (LMCI and EMCI) and Alzheimer's disease (AD).

\paragraph{ASCVD dataset}


We use single nucleotide polymorphisms and metabolomics data from the Emory/Georgia Tech Predictive Health Institute to study the atherosclerosis cardiovascular disease (ASCVD, commonly known as heart disease). The dataset contains 4217 metabolomics features for 236 patients. We select 1000 features out of total 4216 features, which has the largest correlation with the outcome variable: ASCVD risk. The outcome variable is the ASCVD risk score, which measures the risk a patient has to have the disease.

\section{Review of missing data imputation algorithms}\label{ap:alg}
Regarding missing data imputation, various of approaches have been proposed. MICE \cite{van1999flexible} models conditional distribution for each feature and adopts an iterative sequential sampling to learn the true distribution. Following similar scheme, missForest \cite{stekhoven2012missforest} uses random forest to iteratively impute the missing values. More recently, deep generative models are adopted for the imputation task. Denoising autoencoder (DAE) \cite{gondara2018mida, vincent2008extracting} is the first algorithms proposed based on autoencoders. Later \cite{mattei2018miwae} proposes to use importance-weighted autoencoder for missing data imputation with deep latent variable models (DLVM). Gain \cite{yoon2018gain} is the first proposed imputation methods using generative adversarial networks (GAN) \cite{goodfellow2014generative}. It uses generator to impute the missing value and discriminator to specify the distribution of missing values and give feedbacks to generator's imputation. \cite{zhang2018medical} shares similar idea with Gain but uses multiple generators in imputation. Following the idea of Gain, \cite{yang2019categorical} proposes categorical Gain to handle the case with categorical missing features and \cite{friedjungovamissing} proposes Wasserstein-Gain, which further improves the performance when degree of missingness is relatively low. MisGAN \cite{li2019misgan} uses three generators and three discriminators to learn the imputation, which achieves state-of-the-art imputation performances in image recovery. Other approaches such that generative imputation \cite{kachuee2019generative} also shows decent performance in certain applications.

\section{Experiments on ADNI dataset} \label{adni}

We sort the features with descending order in correlation and set the missing length as $L = 100$. Results on imputation fairness are summarized in Table \ref{adni_imp}, and that on prediction fairness are summarized in Table \ref{adni_pred}.

Regarding the imputation fairness, we observe that imputation unfairness appears in the result (Observation 1), and can be associated with imbalances of both missingness (Observation 2) and sample size. In addition, two deep learning imputation models appear to have larger IAPD, even in MCAR mechanisms. Regarding the prediction fairness, we observe that the fairness is also associated with the missing mechanism (Observation 4).

\begin{table*}
\scalebox{0.8}{
\setlength\tabcolsep{1.6pt}
\begin{tabular}{c|cccccccc|cccccccc|cccccccc}
\toprule
     & \multicolumn{8}{c}{MSIE} & \multicolumn{8}{c@{}}{IAPD regarding gender} & \multicolumn{8}{c@{}}{IAPD regarding race}\\
    \midrule
     Method & $I_1$ & $I_2$ & $I_3$ & $I_4$ & $I_5$ & $I_6$ & $I_7$ & Var &
     $I_1$ & $I_2$ & $I_3$ & $I_4$ & $I_5$ & $I_6$ & $I_7$ & VarD$_g$ &
     $I_1$ & $I_2$ & $I_3$ & $I_4$ & $I_5$ & $I_6$ & $I_7$ & VarD$_r$ \\
\midrule
\addlinespace

MCAR (\ref{mcar-1})   
& \textbf{0.36} & 0.46  & 0.61  & 0.42 & 0.71 & 0.74 & 0.87 & 1.00
& 0.08 & \textbf{0.06}  & 0.07  & 0.07 & 0.11 & -0.23 & -0.13 & 0.17 
& \textbf{-0.03} & -0.05  & -0.11  & -0.07 & -0.18 & -0.06 & -0.04 & -0.17 
  \\
MCAR (\ref{mcar-2})   
& \textbf{0.46} & 0.50  & 0.65  & 0.53 & 0.58 & 0.76 & 0.84 & 1.00
& 0.06 & 0.07  & 0.09  & 0.08 & \textbf{0.04} & -0.22 & -0.17 & 0.18
& \textbf{-0.07} & \textbf{-0.07}  & -0.11  & -0.09 & -0.16 & -0.15 & -0.12 & -0.18
  \\

MCAR (\ref{mcar-3})   
& \textbf{0.54} & 0.70  & 0.77  & 0.88 & 1.48 & 0.82 & 0.86 & 1.00
& \textbf{0.06} & \textbf{0.06}  & 0.13  & 0.11 & 0.23 & -0.21 & -0.15 & 0.18
& \textbf{-0.05} & -0.11  & -0.17  & -0.15 & -0.19 & -0.15 & -0.17 & -0.18
  \\ 

\addlinespace
MAR (\ref{mar-1})   
& 0.59 & \textbf{0.58}  & 0.96  & 0.77 & 0.63 & 0.76 & 0.82 & 0.98
& 0.16 & 0.16  & 0.52  & 0.33 & 0.17 & -0.11 & \textbf{-0.08} & 0.18 
& \textbf{-0.04} & -0.05  & -0.24  & -0.20 & -0.13 & -0.27 & -0.25 & -0.31
  \\ 
  
MAR (\ref{mar-2})   
& 0.44 & 0.57  & 0.84  & \textbf{0.38} & 0.59 & 0.89 & 1.11 & 0.86
& \textbf{-0.04} & -0.08  & -0.32  & -0.14 & \textbf{-0.04} & -0.30 & -0.37 & 0.17
& -0.12 & -0.13  & -0.09  & -0.08 & -0.15 & -0.07 & \textbf{0.06} & -0.06
  \\

MAR (\ref{mar-3})   
& 0.57  & \textbf{0.51}  & 0.67  & 0.56 & \textbf{0.51} & 1.17 & 1.11 & 0.92
& \textbf{-0.01} & 0.03  & 0.04  & \textbf{-0.01} & 0.07 & -0.57 & -0.19 & 0.22
& -0.22 & -0.16  & -0.20  & -0.18 & \textbf{-0.07} & -0.21 & -0.43 & -0.32
  \\

MAR (\ref{mar-4})    
& 0.61 & \textbf{0.55} & 0.74 & 0.61 & 0.63 & 0.85 & 0.87 & 0.99
& \textbf{0.07} & 0.11  & 0.17  & 0.18 & 0.11 & \textbf{0.05} & 0.12 & 0.09 
& 0.05 & \textbf{0.01}  & -0.06  & 0.02 & -0.10 & 0.09 & 0.08 & -0.02
  \\

\addlinespace
MNAR (\ref{mnar-1})   
& \textbf{0.63} & 1.97  & 1.94  & 0.96 & 0.68 & 1.18 & 1.16 & 0.35
& -0.19 & -0.64  & -0.67  & -0.14 & \textbf{-0.09} & -0.39 & -0.15 & 0.01
& -0.07 & -0.27  & -0.15  & -0.19 & \textbf{-0.06} & -0.15 & -0.29 & -0.07
  \\

MNAR (\ref{mnar-2})   
& 0.61 & 0.64  & 0.80  & \textbf{0.57} & 0.63 & 1.14 & 1.14 & 0.77
& -0.17 & \textbf{-0.08}  & -0.36  & -0.14 & -0.12 & -0.35 & -0.13 & 0.15
& -0.07 & -0.04  & \textbf{0.01}  & -0.11 & -0.06 & -0.19 & -0.27 & -0.13
  \\

MNAR (\ref{mnar-3})   
& 0.71 & 0.74  & 0.99 & \textbf{0.70} & 0.79 & 0.78 & 0.87 & 0.92
& 0.14 & 0.30  & 0.60  & 0.35 & -0.25 & \textbf{0.07} & 0.09 & 0.09 
& -0.08 & -0.22  & -0.37 & -0.18 & -0.39 & -0.07 & \textbf{-0.02} & -0.14
  \\

MNAR (\ref{mnar-4})   
& \textbf{0.72} & 2.19  & 2.23  & 1.10 & 0.82 & 0.78 & 0.87 & 0.45
& 0.15  & 0.97  & 1.01  & 0.36 & 0.34 & \textbf{0.07} & \textbf{0.07} & 0.11 
& -0.07 & -0.34  & -0.40  & -0.16 & -0.28 & -0.07 & \textbf{-0.02} & -0.08
  \\

\addlinespace
\bottomrule
\end{tabular}
}
\caption{Imputation fairness on ADNI gene dataset. Number of features (besides sensitive attributes) is 1000, $L$ = 100. Here imputation methods are encoded as: $I_1$: MICE; $I_2$: missForest; $I_3$: KNN; $I_4$: SoftImpute; $I_5$: OptSpace; $I_6$: Gain; $I_7$: misGAN. Results are average values over 50 repeated experiments.}
\label{adni_imp}
\end{table*}
\normalsize

\begin{table*}
\scalebox{0.8}{
\setlength\tabcolsep{4.3pt}
\begin{tabular}{c|ccccccc|ccccccc|ccccccc}
\toprule
     & \multicolumn{7}{c}{Prediction Accuracy} & \multicolumn{7}{c@{}}{EOD regarding gender} & \multicolumn{7}{c@{}}{EOD regarding race}\\
    \midrule
     Method & $I_1$ & $I_2$ & $I_3$ & $I_4$ & $I_6$ & $I_7$ & CC &
     $I_1$ & $I_2$ & $I_3$ & $I_4$ & $I_6$ & $I_7$ & CC$_g$ &
     $I_1$ & $I_2$ & $I_3$ & $I_4$ & $I_6$ & $I_7$ & CC$_r$ \\
\midrule
\addlinespace

MCAR (\ref{mcar-1})   
 & 0.33  & 0.33  & 0.33 & 0.33 & 0.30 & 0.33 & 0.33
 & 0.02  & 0.02  & 0.02 & 0.02 & 0.06 & 0.03 & \textbf{0.01}
 & \textbf{0.16}  & 0.20  & \textbf{0.16} & 0.19 & 0.15 & 0.23 & 0.22
  \\
  
MCAR (\ref{mcar-2})   
 & 0.33  & 0.33  & 0.33 & \textbf{0.34} & 0.32 & 0.33 & 0.33
 & 0.02  & \textbf{0.01}  & 0.02 & 0.03 & 0.06 & 0.04 & \textbf{0.01}
 & 0.15  & 0.23  & 0.18 & 0.20 & \textbf{0.14} & 0.22 & \textbf{0.14}
  \\

MCAR (\ref{mcar-3})   
 & 0.33  & 0.33  & 0.32 & 0.33 & 0.30 & 0.33 & 0.33
 & 0.02  & 0.02  & 0.04 & 0.03 & 0.05 & 0.05 & \textbf{0.01}
 & 0.17  & 0.18  & 0.15 & 0.18 & 0.15 & 0.22 & \textbf{0.07}
  \\ 

\addlinespace
MAR (\ref{mar-1})   
 & 0.33  & 0.33  & 0.33 & 0.31  & 0.31 & 0.33 & 0.32
 & \textbf{0.02}  & 0.03  & \textbf{0.02} & 0.07 &  0.06 & 0.05 & 0.05
 & 0.17  & 0.19  & 0.19 & 0.21  & \textbf{0.16} & 0.23 & 0.19
  \\ 
  
MAR (\ref{mar-2})   
 & 0.33  & 0.32  & 0.33 & 0.37 & \textbf{0.31} & 0.33 & 0.33
 & 0.02  & 0.02  & \textbf{0.01} & 0.04  & 0.05 & 0.02 & \textbf{0.01}
 & 0.18  & 0.20  & 0.19 & 0.24 & \textbf{0.17} & 0.21 & 0.18
  \\

MAR (\ref{mar-3})   
 & 0.33  & 0.33  & 0.33 & 0.33 &  0.31 & 0.33 & 0.33
 & \textbf{0.02}  & 0.03  & \textbf{0.02} & 0.03 & 0.05 & 0.06 & 0.03
 & \textbf{0.15}  & 0.20  & 0.21 & 0.18 & 0.16 & 0.26 & 0.25
  \\

MAR (\ref{mar-4})    
 & 0.33 & 0.32  & 0.33 & 0.32 & \textbf{0.31} & 0.33 & 0.33
 & 0.02  & 0.03  & 0.02 & 0.03 & 0.04 & 0.03 & \textbf{0.01}
 & \textbf{0.14}  & 0.21  & 0.20 & 0.16  & 0.15 & 0.18 & 0.16
  \\

\addlinespace
MNAR (\ref{mnar-1})   
 & 0.32 & 0.32  & 0.33 & \textbf{0.34} & 0.31 & 0.33 & 0.33
 & 0.03  & \textbf{0.00}  & \textbf{0.00} & 0.10 &  0.06 & 0.02 & 0.01
 & \textbf{0.13} & 0.18  & 0.19 & 0.27 & 0.17 & 0.17 & 0.22
  \\

MNAR (\ref{mnar-2})   
 & 0.33  & 0.33 & 0.33 & \textbf{0.34} & 0.31 & 0.33 & 0.33
 & 0.02  & 0.02  & 0.02 & 0.04 & 0.04 & 0.04 & \textbf{0.01}
 & 0.17  & 0.23  & 0.18 & 0.17 & \textbf{0.09} & 0.17 & 0.16
  \\

MNAR (\ref{mnar-3})   
 & 0.33  & 0.33 & \textbf{0.34} & 0.33 & 0.31 & 0.33 & 0.33
 & 0.03  & \textbf{0.02}  & 0.03 & 0.05 & 0.06 & 0.04 & \textbf{0.02}
 & 0.15  & 0.21  & 0.21 & 0.20 & \textbf{0.14} & 0.20 & 0.18
  \\

MNAR (\ref{mnar-4})   
 & 0.33  & 0.33  & 0.33 & 0.32 & \textbf{0.31} & 0.33 & 0.32
 & 0.01  & 0.01  & \textbf{0.00} & 0.07 & 0.05 & 0.05 & 0.04
 & 0.14  & 0.14  & 0.17 & 0.24  & \textbf{0.07} & 0.20 & 0.16
  \\

\addlinespace
\bottomrule
\end{tabular}
}
\caption{Prediction fairness on ADNI gene dataset. Number of features (besides sensitive attributes) is 1000, $L$ = 100. Here imputation methods are encoded as: $I_1$: MICE; $I_2$: missForest; $I_3$: KNN; $I_4$: SoftImpute; $I_6$: Gain; $I_7$: misGAN. Due to the experiment setting, in some of the mechanisms, OptSpace implemented in R is not compilable. Hence the method is ignored in this experiment. Results are average values over 50 repeated experiments. When using complete data for prediction, prediction MSE is 0.33, MSED regarding gender is -0.02, MSED regarding race is 0.21. }
\label{adni_pred}
\end{table*}
\normalsize

\begin{table*}
\scalebox{0.8}{
\setlength\tabcolsep{1.6pt}
\begin{tabular}{c|cccccccc|cccccccc|cccccccc}
\toprule
     & \multicolumn{8}{c}{MSIE} & \multicolumn{8}{c@{}}{IAPD regarding gender} & \multicolumn{8}{c@{}}{IAPD regarding race}\\
    \midrule
     Method & $I_1$ & $I_2$ & $I_3$ & $I_4$ & $I_5$ & $I_6$ & $I_7$ & Var &
     $I_1$ & $I_2$ & $I_3$ & $I_4$ & $I_5$ & $I_6$ & $I_7$ & VarD$_g$ &
     $I_1$ & $I_2$ & $I_3$ & $I_4$ & $I_5$ & $I_6$ & $I_7$ & VarD$_r$ \\
\midrule
\addlinespace

MCAR (\ref{mcar-1})   
& 0.65 & \textbf{0.64}  & 0.79  & 0.67 & 0.71 & 0.82 & 1.25 & 1.00
& 0.01 & \textbf{0.00}  & 0.03  & \textbf{0.00} & -0.01 & 0.06 & -0.06 & -0.03 
& \textbf{-0.07} & -0.07  & -0.13  & -0.10 & -0.09 & -0.11 & -0.18 & -0.21 
  \\
MCAR (\ref{mcar-2})   
& \textbf{0.64} & 0.69  & 0.85  & 0.79 & 0.74 & 0.85 & 1.07 & 1.00
& \textbf{0.00} & -0.01  & 0.02  & -0.02 & -0.01 & 0.06 & -0.00 & -0.04
& \textbf{-0.07} & -0.10  & -0.17  & -0.16 & -0.12 & -0.17 & -0.27 & -0.23
  \\

MCAR (\ref{mcar-3})   
& \textbf{0.75} & 0.92  & 0.99  & 0.95 & 0.82 & 0.93 & 1.01 & 1.00
& \textbf{0.01} & -0.02  & \textbf{0.01}  & -0.03 & -0.02 & 0.05 & -0.02 & 0.18
& \textbf{-0.11} & -0.17  & -0.22  & -0.23 & -0.16 & -0.14 & -0.24 & -0.24
  \\ 

\addlinespace
MAR (\ref{mar-1})   
& \textbf{0.65} & 0.69 & 0.92  & 0.78 & 0.74 & 0.88 & 1.11 & 0.97
& 0.02 & 0.04  & 0.10  & 0.08 & \textbf{0.01} & 0.08 & -0.07 & -0.03 
& \textbf{-0.07} & -0.13  & -0.17  & -0.11 & -0.20 & \textbf{-0.07} & -0.15 & -0.19
  \\ 
  
MAR (\ref{mar-2})   
& \textbf{0.66} & 0.76  & 0.90  & 0.87 & 0.83 & 0.87 & 1.04 & 1.00
& -0.05 & -0.07  & -0.08  & -0.11 & \textbf{-0.02} & 0.04 & \textbf{0.02} & -0.04
& \textbf{-0.06} & -0.11  & -0.21  & -0.22 & -0.15 & -0.16 & 0.28 & -0.15
  \\

MAR (\ref{mar-3})   
& 0.78  & \textbf{0.77}  & 0.95  & 0.86 & 0.79 & 1.17 & 1.28 & 1.05
& -0.09 & \textbf{-0.01}  & 0.03  & -0.02 & 0.07 & -0.09 & -0.12 & -0.05
& -0.21 & \textbf{-0.18}  & -0.28  & -0.26 & -0.20 & -0.52 & -0.68 & -0.35
  \\

MAR (\ref{mar-4})    
& \textbf{0.63} & 0.67 & 0.82 & 0.74 & 0.72 & 0.74 & 0.85 & 0.93
& 0.03 & -0.02  & \textbf{-0.01}  & -0.03 & \textbf{-0.01} & -0.09 & -0.14 & 0.04
& \textbf{-0.03} & -0.06  & -0.12  & -0.11 & -0.08 & -0.06 & -0.08 & -0.14
  \\

\addlinespace
MNAR (\ref{mnar-1})   
& \textbf{0.80} & 2.27  & 2.07  & 1.10 & 1.29 & 1.22 & 1.31 & 0.52
& -0.13 & -0.16  & -0.05  & \textbf{-0.02} & -0.31 & -0.22 & -0.21 & -0.02
& \textbf{-0.17} & -0.28  & -0.22  & -0.23 & -0.79 & -0.70 & -0.88 & -0.16
  \\

MNAR (\ref{mnar-2})   
& \textbf{0.80} & 1.04  & 1.18  & 0.89 & 0.85 & 1.19 & 1.33 & 0.90
& -0.14 & -0.04  & 0.04  & \textbf{-0.01} & -0.10 & -0.15 & -0.22 & -0.06
& -0.16 & \textbf{-0.14}  & -0.15  & -0.17 & -0.26 & -0.58 & -0.91 & -0.21
  \\

MNAR (\ref{mnar-3})   
& \textbf{0.67} & 0.96  & 0.94 & 0.82 & 0.77 & 0.80 & 0.89 & 0.82
& 0.11 & \textbf{0.01}  & -0.03  & -0.05 & 0.03 & 0.03 & 0.06 & -0.01
& \textbf{0.03} & 0.26  & -0.30 & -0.27 & -0.19 & -0.05 & -0.08 & -0.18
  \\

MNAR (\ref{mnar-4})   
& \textbf{0.67} & 2.06  & 1.80  & 1.00 & 1.01 & 0.80 & 0.88 & 0.44
& 0.11  & -0.08  & -0.09  & -0.05 & 0.06 & \textbf{0.04} & 0.05 & -0.01
& \textbf{0.04} & -0.42  & -0.39  & -0.28 & -0.28 & \textbf{-0.04} & -0.06 & -0.16
  \\

\addlinespace
\bottomrule
\end{tabular}
}
\caption{Imputation fairness on ASCVD dataset. Number of features (besides sensitive attributes) is 1000, $L$ = 100. Here imputation methods are encoded as: $I_1$: MICE; $I_2$: missForest; $I_3$: KNN; $I_4$: SoftImpute; $I_5$: OptSpace; $I_6$: Gain; $I_7$: misGAN. Results are average values over 50 repeated experiments.}
\label{cvd_imp}
\end{table*}
\normalsize

\begin{table*}[t]
\scalebox{0.8}{
\setlength\tabcolsep{1.9pt}
\begin{tabular}{c|cccccccc|cccccccc|cccccccc}
\toprule
     & \multicolumn{8}{c}{Prediction MSE} & \multicolumn{8}{c@{}}{RED regarding gender} & \multicolumn{8}{c@{}}{RED regarding race}\\
    \midrule
     Method & $I_1$ & $I_2$ & $I_3$ & $I_4$ & $I_5$ & $I_6$ & $I_7$ & CC &
     $I_1$ & $I_2$ & $I_3$ & $I_4$ & $I_5$ & $I_6$ & $I_7$ & CC$_g$ &
     $I_1$ & $I_2$ & $I_3$ & $I_4$ & $I_5$ & $I_6$ & $I_7$ & CC$_r$ \\
\midrule
\addlinespace

MCAR (\ref{mcar-1})   
 & \textbf{1.41}  & 1.50  & 1.49 & 2.56 & 2.12 & 2.22 & 1.50 & 1.49
 & 0.42  & 0.20  & 0.24 & -0.27 & 0.34 & 0.56 & \textbf{0.13} & 0.23
 & \textbf{-0.70}  & -1.49  & -1.46 & -0.85 & -2.72 & -3.14 & -1.34 & -1.52
  \\
  
MCAR (\ref{mcar-2})   
 & \textbf{1.43}  & 1.51  & 1.49 & 2.58 & 2.19 & 2.22 & 1.52 & 1.52
 & 0.51  & 0.20  & 0.19 & -0.23 & 0.31 & 0.56 & \textbf{0.13} & 0.22
 & \textbf{-0.54}  & -1.42  & -1.38 & -0.82 & -2.89 & -3.13 & -0.46 & -1.60
  \\

MCAR (\ref{mcar-3})   
 & 1.89  & \textbf{1.63}  & 1.92 & 3.03 & 2.21 & 2.21 & 1.87 & 1.71
 & 0.31  & 0.07  & 0.05 & -0.18 & 0.43 & 0.55 & \textbf{-0.04} & 0.13
 & \textbf{-0.34}  & -1.37  & -1.14 & -0.68 & -2.76 & -3.13 & 1.30 & -1.97
  \\ 

\addlinespace
MAR (\ref{mar-1})   
 & \textbf{1.38}  & 1.54  & 1.54 & 2.71 & 2.07 & 2.21 & 1.53 & 1.51
 & 0.51  & 0.16  & 0.17 & -0.27 & 0.33 & 0.55 & \textbf{0.10} & 0.20
 & \textbf{-0.72}  & -1.65  & -1.66 & -0.93 & -2.68 & -3.13 & -0.49 & -1.90
  \\ 
  
MAR (\ref{mar-2})   
 & \textbf{1.43}  & 1.52  & 1.52 & 2.59 & 2.19 & 2.22 & 1.56 & 1.62
 & 0.49  & 0.24  & 0.21 & -0.31 & 0.32 & 0.56 & 0.19 & \textbf{0.18}
 & \textbf{-0.49}  & -1.30  & -1.30 & -0.77 & -2.85 & -3.13 & -1.32 & -1.38
  \\

MAR (\ref{mar-3})   
 & \textbf{1.36}  & 1.45  & 1.46 & 2.48 & 2.21 & 2.21 & 1.54 & 1.55
 & 0.37  & 0.18 & 0.18 & -0.21 & 0.43 & 0.55 & \textbf{0.11} & 0.29
 & \textbf{-0.45}  & -1.17  & -1.11 & -0.64 & -2.78 & -3.13 & -1.26 & -1.87
  \\

MAR (\ref{mar-4})    
 & \textbf{1.45} & 1.57 & 1.58  & 2.63 & 2.11 & 2.21 & 1.62 & 1.54
 & 0.41 & 0.15 & 0.14 & -0.22 & 0.33 & 0.55 & \textbf{0.03} & 0.18
 & \textbf{-0.77}  & -1.59  & -1.59 & -0.92 & -2.67 & -3.13 & -1.57 & -1.56
  \\

\addlinespace
MNAR (\ref{mnar-1})   
 & 1.72 & 1.82  & 1.85 & 2.73 & 1.85 & 2.22 & 1.69 & \textbf{1.62}
 & 0.35  & 0.07  & 0.14 & -0.21 & \textbf{-0.04} & 0.55 & 0.31 & 0.06
 & \textbf{-0.49} & -1.17  & -1.09 & -0.71 & -1.88 & -3.14 & -1.09 & -1.57
  \\

MNAR (\ref{mnar-2})   
 & \textbf{1.44}  & 1.50  & 1.49 & 2.41 & 1.96 & 2.22 & 1.66 & 1.55
 & 0.51 & 0.26  & 0.24 & \textbf{0.10} & 0.11 & 0.55 & 0.31 & 0.19
 & \textbf{-0.65}  & -1.34  & -1.33 & 0.71 & -2.35 & -3.13 & -1.08 & -1.53
  \\

MNAR (\ref{mnar-3})   
 & \textbf{1.51}  & 1.57 & 1.58 & 3.02 & 2.22 & 2.21 & 1.66 & 1.53
 & 0.39  & 0.12  & 0.13 & 0.34 & 0.31 & 0.55 & \textbf{0.03} & 0.21
 & \textbf{-0.50}  & -1.55  & -1.52 & -0.66 & -2.84 & -3.13 & -1.72 & -1.78
  \\

MNAR (\ref{mnar-4})   
 & 2.97  & 2.38  & 2.39 & 3.42 & 2.36 & 2.22 & 1.72 & \textbf{1.60}
 & 0.19  & -0.62  & -0.61 & -0.51 & -0.20 & 0.55 & \textbf{-0.02} & 0.20
 & \textbf{-0.75} & -1.48  & -1.49 & -0.79 & -2.63 & -3.13 & -1.80 & -2.15
  \\

\addlinespace
\bottomrule
\end{tabular}
}
\caption{Prediction fairness on ASCVD dataset. Number of features (besides sensitive attributes) is 1000, $L$ = 100. Here imputation methods are encoded as: $I_1$: MICE; $I_2$: missForest; $I_3$: KNN; $I_4$: SoftImpute; $I_5$: OptSpace; $I_6$: Gain; $I_7$: misGAN. Results are average values over 50 repeated experiments. When using complete data for prediction, prediction MSE is 1.49, MSED regarding gender is 0.23, MSED regarding race is -1.46.}
\label{cvd_pred}
\end{table*}
\normalsize

\section{Experiments on ASCVD dataset} \label{cvd}
We sort the features with descending order in correlation and set the missing length as $L = 100$. Results on imputation fairness are summarized in Table \ref{cvd_imp}. Regarding prediction fairness, since the outcome $y$ is continuous, we define \textit{regression error difference} as the fairness notion:
\begin{definition}[regression error difference]
Regression error difference for regression algorithm $h$ is defined as $\text{RED}(h) := \text{MSE}_{\text{maj}}(h) - \text{MSE}_{\text{min}}(h)$
\end{definition}
The experiment results are summarized in Table \ref{cvd_pred}. Again, regarding the imputation fairness, we observe the unfairness of imputation (Observation 1), which appears to be associated with imbalances of both missingness (Observation 2) and sample size. In addition, we observe that self-contribution in the propensity score model also has significant impact on imputation unfairness. Two missing data mechanisms (3a) and (3d) are only different in the part related to self-contribution from missing values. We observe that imputation unfairness often differs much between these two mechanisms for among all imputation models. Regarding the prediction fairness, we also observe that the prediction fairness is associated with the missing mechanism (Observation 4).

\section{Experiment details}
In all the experiments, Gain (implemented), Misgan (implemented, original code is designed for image data) and MICE (using \texttt{IterativeImputer} in Python package \texttt{scikit-learn}) are run in Pytorch (version 3.6.9) in Google Colab, other methods are run in R (version 3.6.1). Missforest is run using R package \texttt{missForest}. KNN, SoftImpute and OptSpace are run using R package \texttt{filling}. To ensure that the random seeds are the same, we use python package \texttt{rpy2} to use the R code in generating the data (i.e., random train-test split) and training the predition model (random forest, using R package \texttt{randomForest}).

\end{document}